\documentclass{sig-alternate-05-2015}
\usepackage{caption}
\begin{document}

\setcopyright{acmcopyright}




\acmPrice{\$15.00}

%
\conferenceinfo{SIGIR}{2016 Pisa, Italy}

\title{Tweet2Vec: Learning Tweet Embeddings Using Character-level CNN-LSTM Encoder-Decoder}

%
%
%
%
%

\numberofauthors{3} 
%
\author{
\alignauthor
Soroush Vosoughi\thanks{The first two authors contributed equally to this work.}\\
\affaddr{MIT Media Lab}\\
\email{soroush@mit.edu}
\alignauthor 
\mbox{Prashanth Vijayaraghavan\footnotemark[1]}\\
\affaddr{MIT Media Lab}\\
\email{pralav@mit.edu}
\and
\alignauthor 
Deb Roy\\
\affaddr{MIT Media Lab}\\
\email{dkroy@media.mit.edu}
}

\date{30 July 1999}

\CopyrightYear{2016} 
\setcopyright{acmlicensed}
\conferenceinfo{SIGIR '16,}{July 17 - 21, 2016, Pisa, Italy}
\isbn{978-1-4503-4069-4/16/07}\acmPrice{\$15.00}
\doi{http://dx.doi.org/10.1145/2911451.2914762}

\maketitle
\begin{abstract}
We present \emph{Tweet2Vec}, a novel method for generating general-purpose vector representation of tweets. The model learns tweet embeddings using character-level CNN-LSTM encoder-decoder. We trained our model on 3 million, randomly selected English-language tweets. The model was evaluated using two methods: tweet semantic similarity and tweet sentiment categorization, outperforming the previous state-of-the-art in both tasks. The evaluations demonstrate the power of the tweet embeddings generated by our model for various tweet categorization tasks. The vector representations generated by our model are generic, and hence can be applied to a variety of tasks. Though the model presented in this paper is trained on English-language tweets, the method presented can be used to learn tweet embeddings for different languages.
\end{abstract}

%
%


%
%

%
%

\linespread{.91}


\setlength{\belowcaptionskip}{-5pt} 

\section{Introduction}
In recent years, the micro-blogging site Twitter has become a major social media platform with hundreds of millions of users. The short (140 character limit), noisy and idiosyncratic nature of tweets make standard information retrieval and data mining methods ill-suited to Twitter. Consequently, there has been an ever growing body of IR and data mining literature focusing on Twitter. 
However, most of these works employ extensive feature engineering to create task-specific, hand-crafted features. This is time consuming and inefficient as new features need to be engineered for every task.

In this paper, we present \emph{Tweet2Vec}, a method for generating general-purpose vector representation of tweets that can be used for any classification task. \emph{Tweet2Vec} removes the need for expansive feature engineering and can be used to train any standard off-the-shelf classifier (e.g., logistic regression, svm, etc). \emph{Tweet2Vec} uses a CNN-LSTM encoder-decoder model that operates at the character level to learn and generate vector representation of tweets. Our method is especially useful for natural language processing tasks on Twitter where it is particularly difficult to engineer features, such as speech-act classification and stance detection (as shown in our previous works on these topics \cite{vosoughi_act_2016,vsvr_stance}). 

There has been several works on generating embeddings for words, most famously \emph{Word2Vec} by Mikolov et al. \cite{mikolov2013distributed}). There has also been a number of different works that use encoder-decoder models based on long short-term memory (LSTM) \cite{sutskever2014sequence}, and gated recurrent neural networks (GRU) \cite{chung2014empirical}. These methods have been used mostly in the context of machine translation. The encoder maps the sentence from the source language to a vector representation, while the decoder conditions on this encoded vector for translating it to the target language. Perhaps the work most related to ours is the work of Le and Mikolov \shortcite{le2014distributed}, where they extended the \emph{Word2Vec} model to generate representations for sentences (called \emph{ParagraphVec}). However, these models all function at the word level, making them ill-suited to the extremely noisy and idiosyncratic nature of tweets. Our character-level model, on the other hand, can better deal with the noise and idiosyncrasies in tweets. We plan to make our model and the data used to train it publicly available to be used by other researchers that work with tweets.

\section{CNN-LSTM Encoder-Decoder}
In this section, we describe the CNN-LSTM encoder-decoder model that operates at the character level and generates vector representation of tweets. The encoder consists of convolutional layers to extract features from the characters and an LSTM layer to encode the sequence of features to a vector representation, while the decoder consists of two LSTM layers which predict the character at each time step from the output of encoder.

\subsection{Character-Level CNN Tweet Model}\label{cnn_char}
Character-level CNN (CharCNN) is a slight variant of the deep character-level convolutional neural network introduced by Zhang et al \cite{zhang2015text}. In this model, we perform temporal convolutional and temporal max-pooling operations, which computes one-dimensional convolution and pooling functions, respectively, between input and output. Given a discrete input function $f(x)\in [1,l] \mapsto \mathbb{R}$, a discrete kernel function $k(x) \in [1,m] \mapsto \mathbb{R}$ and stride $s$, the convolution $g(y)\in [1,(l-m+1)/s] \mapsto \mathbb{R}$  between $k(x)$ and $f(x)$ and pooling operation $h(y) \in [1,(l-m+1)/s] \mapsto \mathbb{R}$ of $f(x)$ is calculated as:
\begin{equation}
g(y)=\sum_{x=1}^{m}{k(x)\cdot f(y\cdot s-x+c)}
\end{equation}

\begin{equation}
h(y)=max_{x=1}^{m}{f(y\cdot s-x+c)}
\end{equation}
where $c=m-s+1$ is an offset constant.

We adapted this model, which employs temporal convolution and pooling operations, for tweets. The character set includes the English alphabets, numbers, special characters and unknown character. There are 70 characters in total, given below:\\

 \indent\indent\indent\indent\texttt{abcdefghijklmnopqrstuvwxyz0123456789}\\
\indent\indent\indent\indent \texttt{-,;.!?:'"/\textbackslash|\_\@\#\$\%\&\char`\^ *\textasciitilde`+-=<>()[]\{\}}\\

Each character in the tweets can be encoded using one-hot vector $x_i\in\{0,1\}^{70}$. Hence, the tweets are represented as a binary matrix $x_{1..150}\in\{0,1\}^{150x70}$ with padding wherever necessary, where 150 is the maximum number of characters in a tweet (140 tweet characters and padding) and 70 is the size of the character set. 

Each tweet, in the form of a matrix, is now fed into a deep model consisting of four 1-d convolutional layers. A convolution operation employs a filter $w \in \mathbb{R}^l$, to extract n-gram character feature from a sliding window of $l$ characters at the first layer and learns abstract textual features in the subsequent layers. The convolution in the first layer operates on sliding windows of character (size $l$), and the convolutions in deeper layers are defined in a similar way. Generally, for tweet $s$, a feature $c_i$ at layer $h$ is generated by:

\begin{equation}\label{conv}c_i^{(h)}(s)=g(w^{(h)}.\hat{c}_i^{(h-1)}+b^{(h)})\end{equation}
where 
$\hat{c}_i^{(0)}=x_{i...i+l-1}$, $b^{(h)} \in \mathbb{R}$ is the bias at layer $h$ and $g$ is a rectified linear unit.

This filter $w$ is applied across all possible windows of characters in the tweet to produce a feature map. The output of the convolutional layer is followed by a 1-d max-overtime pooling operation \cite{collobert2011natural} over the feature map and selects the maximum value as the prominent feature from the current filter. In this way, we apply $n$ filters at each layer. Pooling size may vary at each layer (given by $p^{(h)}$ at layer $h$). The pooling operation shrinks the size of the feature representation and filters out trivial features like unnecessary combination of characters. The window length $l$, number of filters $f$, pooling size $p$ at each layer are given in Table \ref{layers}.

\linespread{.93}
\begin{table}[h!]%
\small
\centering
\caption{Layer Parameters of CharCNN}
\label{layers} %
\begin{tabular}{ |c|c|c|c| }
\hline %
$ Layer$  &$Window$  & $Filters$ & $Pool$  \\
($h$)&  $Size$ ($l$) & ($f$)& $Size$ ($p$) \\\hline %
1   & 7 & 512 & 3 \\\hline
2  &7 &512 & 3 \\\hline
3&  3 & 512 & N/A\\\hline
4  &3 & 512 & N/A \\\hline
\end{tabular}
\end{table}
\linespread{.91}

We define $CharCNN(T)$ to denote the character-level CNN operation on input tweet matrix $T$.
The output from the last convolutional layer of CharCNN(T) (size: $10\times 512$) is subsequently given as input to the LSTM layer. Since LSTM works on sequences (explained in Section \ref{lstm} and \ref{model}), pooling operation is restricted to the first two layers of the model (as shown in Table \ref{layers}).

\begin{figure}[ht]
  \includegraphics[width=.92\columnwidth]{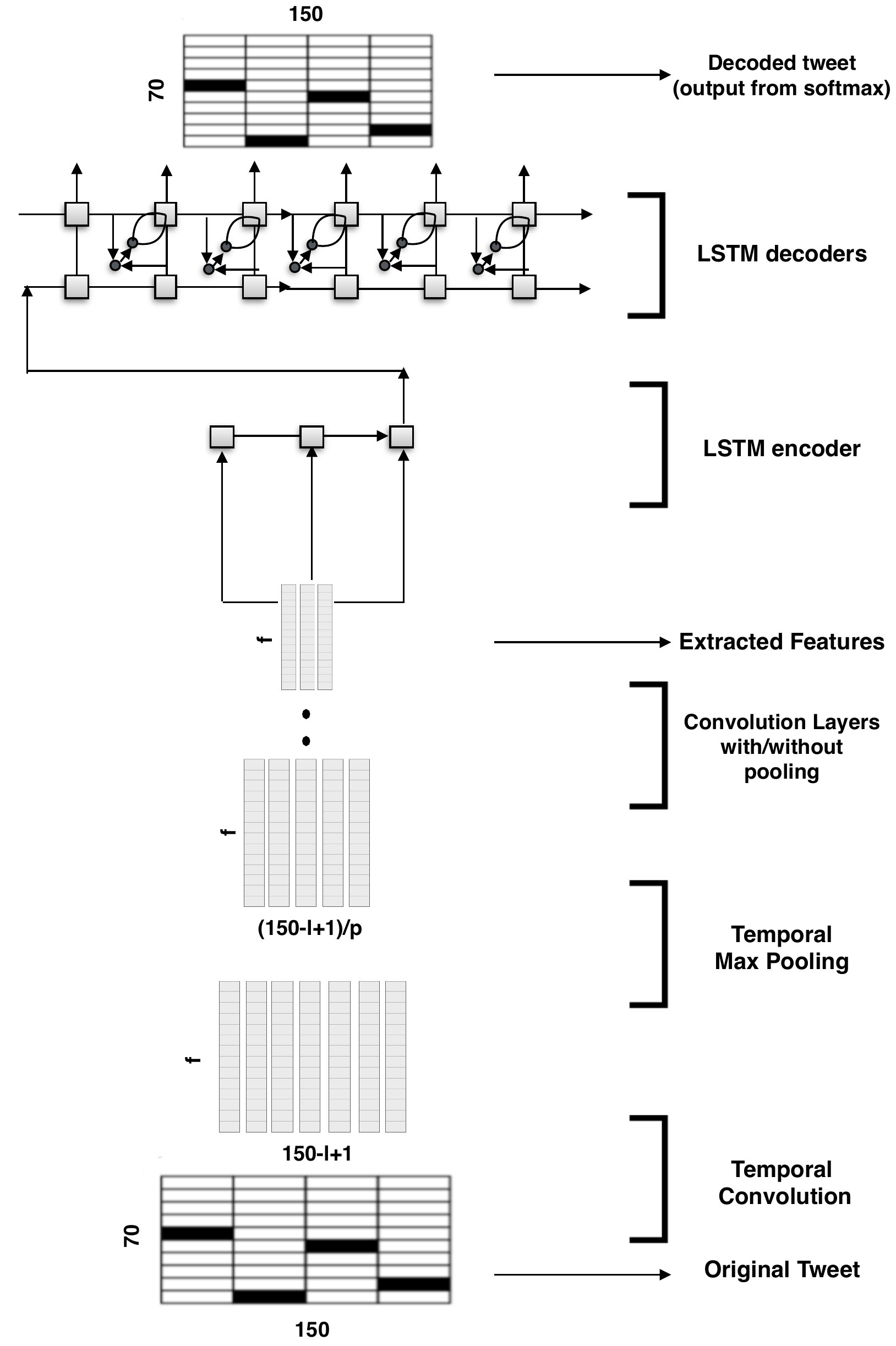}
  \caption{Illustration of the CNN-LSTM Encoder-Decoder Model}\label{comp}
\end{figure}

\subsection{Long-Short Term Memory (LSTM)}\label{lstm}
In this section we briefly describe the LSTM model \cite{hochreiter1997long}. Given an input sequence $X=$($x_1, x_2, ..., x_N$), LSTM computes the hidden vector sequence $h=$($h_1, h_2, ..., h_N$) and and output vector sequence $Y=$($y_1, y_2, ..., y_N$). At each time step, the output of the module is controlled by a set of gates as a function of the previous hidden state $h_{t−1}$ and the input at the current time step $x_t$, the forget gate $f_t$, the input gate $i_t$, and the output gate $o_t$. These gates collectively decide the transitions of the current memory cell $c_t$ and the current hidden state $h_t$. The LSTM transition functions are defined as follows:
\begin{equation}
\begin{split}
i_t = \sigma(W_i \cdot [h_{t-1}, x_t] + b_i)\\
f_t = \sigma(W_f \cdot [h_{t-1}, x_t] + b_f )\\
l_t = tanh(W_l \cdot [h_{t-1}, x_t] + b_l)\\
o_t = \sigma(W_o · [h_{t-1}, x_t] + b_o)\\
c_t = f_t \odot c_{t-1} + i_t \odot l_t\\
h_t = o_t \odot tanh(c_t)
\end{split}
\end{equation}

Here, $\sigma$ is the $sigmoid$ function that has an output in [0, 1], $tanh$ denotes the hyperbolic tangent function that has an output in $[-1, 1]$, and $\odot$ denotes the component-wise multiplication. The extent to which the information in the old memory cell is discarded is controlled by $f_t$, while $i_t$  controls the extent to which new information is stored in the current memory cell, and $o_t$ is the output based on the memory cell $c_t$. LSTM is explicitly designed for learning long-term dependencies, and therefore we choose LSTM after the convolution layer to learn dependencies in the sequence of extracted features.
In sequence-to-sequence generation tasks, an LSTM defines a distribution over outputs and sequentially predicts tokens using a softmax function.
\begin{equation}
P(Y|X)=\prod_{t\in [1,N]}\frac{exp(g(h_{t-1}, y_t))}{\sum_{y'}{exp(g(h_{t−1}, y'_t})}
\end{equation}
where $g$ is the activation function.
For simplicity, we define $LSTM(x_t,h_{t-1})$ to denote the LSTM operation on input $x$ at time-step $t$ and the previous hidden state $h_{t-1}$.

\subsection{The Combined Model}\label{model}
The CNN-LSTM encoder-decoder model draws on the intuition that the sequence of features (e.g. character and word n-grams) extracted from CNN can be encoded into a vector representation using LSTM that can embed the meaning of the whole tweet. Figure \ref{comp} illustrates the complete encoder-decoder model. The input and output to the model are the tweet represented as a matrix where each row is the one-hot vector representation of the characters. The procedure for encoding and decoding is explained in the following section.

\subsubsection{Encoder}
Given a tweet in the  matrix form T (size: $150\times70$), the CNN (Section \ref{cnn_char}) extracts the features from the character representation. The one-dimensional convolution involves a filter vector sliding over a sequence and detecting features at different positions. The new successive higher-order window representations then are fed into LSTM (Section \ref{lstm}). Since LSTM extracts representation from sequence input, we will not apply pooling after convolution at the higher layers of Character-level CNN model. The encoding procedure can be summarized as:
\begin{equation}\label{conv_eq}
H^{conv}=CharCNN(T)   
\end{equation}
\begin{equation}\label{lstm_eq}
h_t=LSTM(g_t, h_{t-1})
\end{equation}
where $g=H^{conv}$ is an extracted feature matrix where each row can be considered as a time-step for the LSTM and $h_t$ is the hidden representation at time-step $t$. LSTM operates on each row of the $H^{conv}$ along with the hidden vectors from previous time-step to produce embedding for the subsequent time-steps. The vector output at the final time-step, $enc_N$, is used to represent the entire tweet. In our case, the size of the $enc_N$ is 256.

\subsubsection{Decoder}
The decoder operates on the encoded representation with two
layers of LSTMs. In the initial time-step, the end-to-end output from the encoding procedure is used as the original input into first LSTM layer. The last LSTM decoder generates each character, $C$, sequentially and combines it with previously generated hidden vectors of size 128, $h_{t-1}$, for the next time-step prediction. The prediction of character at each time step is given by:

\begin{equation}
P(C_t | \cdot) = softmax(T_t, h_{t-1}) 
\end{equation}
where $C_t$ refers to the character at time-step $t$, $T_t$ represents the one-hot vector of the character at time-step $t$. The result from the softmax is a decoded tweet matrix $T^{dec}$, which is eventually compared with the actual tweet or a synonym-replaced version of the tweet (explained in Section \ref{data_aug}) for learning the parameters of the model.


\section{Data Augmentation \& Training}\label{data_aug}
We trained the CNN-LSTM encoder-decoder model on 3 million randomly selected English-language tweets populated using data augmentation techniques, which are useful for controlling generalization error for deep learning models. Data augmentation, in our context, refers to replicating tweet and replacing some of the words in the replicated tweets with their synonyms. These synonyms are obtained from WordNet \cite{fellbaum1998wordnet} which contains words grouped together on the basis of their meanings. This involves selection of replaceable words (example of non-replaceable words are stopwords, user names, hash tags, etc) from the tweet and the number of words $n$ to be replaced. The probability of the number, $n$, is given by a geometric distribution with parameter $l$ in which $P[n] \sim l^n$. Words generally have several synonyms, thus the synonym index $m$, of a given word is also determined by another geometric distribution in which $P[s] \sim r^m$. In our encoder-decoder model, we decode the encoded representation to the actual tweet or a synonym-replaced version of the tweet from the augmented data. We used $p=0.5$, $r=0.5$ for our training. We also make sure that the POS tags of the replaced words are not completely different from the actual words. For regularization, we apply a dropout mechanism after the penultimate layer. This prevents co-adaptation of hidden units by randomly setting a proportion $\rho$ of the hidden units to zero (for our case, we set $\rho=0.5$).

To learn the model parameters, we minimize the cross-entropy loss as the training objective using the Adam Optimization algorithm \cite{kingma2014adam}. It is given by
\begin{equation}\label{cat_ce} CrossEnt(p,q)=-\sum p(x)\log(q(x))\end{equation}
where p is the true distribution (one-hot vector representing characters in the tweet) and q is the output of the softmax. This, in turn, corresponds to computing the negative log-probability of the true class.

\section{Experiments}
We evaluated our model using two classification tasks: \emph{Tweet semantic relatedness} and \emph{Tweet sentiment classification}. 

\subsection{Semantic Relatedness}
The first evaluation is based on the SemEval 2015-Task 1: \emph{Paraphrase and Semantic Similarity in Twitter} \cite{xu2015semeval}. Given a pair of tweets, the goal is to predict their semantic equivalence (i.e., if they express the same or very similar meaning), through a binary yes/no judgement. The dataset provided for this task contains 18K tweet pairs for training and 1K pairs for testing, with $35\%$ of these pairs being paraphrases, and $65\%$ non-paraphrases. 

We first extract the vector representation of all the tweets in the dataset using our \emph{Tweet2Vec} model. We use two features to represent a tweet pair. Given two tweet vectors $r$ and $s$, we compute their element-wise product $r \cdot s$ and their absolute difference $|r - s|$ and concatenate them together (Similar to \cite{kiros2015skip}). We then train a logistic regression model on these features using the dataset. Cross-validation is used for tuning the threshold for classification. In contrast to our model, most of the methods used for this task were largely based on extensive use of feature engineering, or a combination of feature engineering with semantic spaces. Table 2 shows the performance of our model compared to the top four models in the SemEval 2015 competition, and also a model that was trained using \emph{ParagraphVec}. Our model (\emph{Tweet2Vec}) outperforms all these models, without resorting to extensive task-specific feature engineering.

\linespread{.93}
\begin{table}[ht]%
\centering
\small
\label{results_semeval} %
\caption{Results of the paraphrase and semantic similarity in Twitter task.}

\begin{tabular}{ |c|c|c|c| }
\hline %
$ Model$  &$Precision$  & $Recall$ & $F1-Score$  \\\hline %
ParagraphVec & 0.570 & 0.680 & 0.620 \\\hline
nnfeats &  \textbf{0.767} &0.583&  0.662\\\hline
ikr & 0.569 & \textbf{0.806} &0.667 \\\hline
linearsvm & 0.683 & 0.663 &0.672 \\\hline
svckernel   &   0.680& 0.669   & 0.674 \\\hline
\textbf{Tweet2Vec}  &0.679 &  0.686 & \textbf{0.677} \\\hline
\end{tabular}
\end{table}
\linespread{.91}

\subsection{Sentiment Classification}
The second evaluation is based on the SemEval 2015-Task 10B: \emph{Twitter Message Polarity Classification} \cite{rosenthal-EtAl:2015:SemEval}. Given a tweet, the task is to classify it as either positive, negative or neutral in sentiment. The size of the training and test sets were 9,520 tweets and 2,380 tweets respectively ($38\%$ positive, $15\%$ negative, and  $47\%$ neutral).

As with the last task, we first extract the vector representation of all the tweets in the dataset using \emph{Tweet2Vec} and use that to train a logistic regression classifier using the vector representations. Even though there are three classes, the SemEval task is a binary task. The performance is measured as the average F1-score of the positive and the negative class. Table 3 shows the performance of our model compared to the top four models in the SemEval 2015 competition (note that only the F1-score is reported by SemEval for this task) and \emph{ParagraphVec}. Our model outperforms all these models, again without resorting to any feature engineering. 

\linespread{.93}
\begin{table}[ht]%
\centering
\small
\label{results_senti} %
\caption{Results of Twitter sentiment classification task.}
\begin{tabular}{ |c|c|c|c| }
\hline %
$ Model$  &$Precision$  & $Recall$ & $F1-Score$  \\\hline %
ParagraphVec & 0.600 & 0.680 & 0.637 \\\hline
INESC-ID & N/A & N/A & 0.642 \\\hline
lsislif   &   N/A &  N/A   & 0.643 \\\hline
unitn &  N/A &  N/A &0.646 \\\hline
Webis &   N/A &  N/A &  0.648\\\hline
\textbf{Tweet2Vec}  & 0.675 &  0.719 &  \textbf{0.656} \\\hline
\end{tabular}
\end{table}
\linespread{.91}

\vspace{-2mm}
\section{Conclusion and Future Work}
In this paper, we presented \emph{Tweet2Vec}, a novel method for generating general-purpose vector representation of tweets, using a character-level CNN-LSTM encoder-decoder architecture. To the best of our knowledge, ours is the first attempt at learning and applying character-level tweet embeddings. Our character-level model can deal with the noisy and peculiar nature of tweets better than methods that generate embeddings at the word level. Our model is also robust to synonyms with the help of our data augmentation technique using WordNet.

The vector representations generated by our model are generic, and thus can be applied to tasks of different nature. We evaluated our model using two different SemEval 2015 tasks: Twitter semantic relatedness, and sentiment classification. 
Simple, off-the-shelf logistic regression classifiers trained using the vector representations generated by our model outperformed the top-performing methods for both tasks, without the need for any extensive feature engineering. This was despite the fact that due to resource limitations, our \emph{Tweet2Vec} model was trained on a relatively small set (3 million tweets). Also, our method outperformed \emph{ParagraphVec}, which is an extension of \emph{Word2Vec} to handle sentences. This is a small but noteworthy illustration of why our tweet embeddings are best-suited to deal with the noise and idiosyncrasies of tweets.

For future work, we plan to extend the method to include: 1) Augmentation of data through reordering the words in the tweets to make the model robust to word-order, 2) Exploiting attention mechanism \cite{li2015hierarchical} in our model to improve alignment of words in tweets during decoding, which could improve the overall performance.



%

\bibliographystyle{abbrv}
\bibliography{tweet2vec}  
%
%
\end{document}